\documentclass[conference]{IEEEtran}
\ifCLASSINFOpdf
\else
\fi
\usepackage{amsmath,amsfonts,amssymb}
\usepackage{graphicx}
\usepackage[colorlinks=true, allcolors=blue]{hyperref}
\usepackage[font=footnotesize]{subfig}
\usepackage{threeparttable}
\usepackage{multirow}

\begin{document}
%
\title{A CNN Based Scene Chinese Text Recognition Algorithm With Synthetic Data Engine}

\author{\IEEEauthorblockN{Xiaohang Ren, Kai Chen and Jun Sun}
\IEEEauthorblockA{Institute of Image Communication and Network Engineering,
Shanghai Jiao Tong University, Shanghai, China.\\
Email: [xiaomu, kchen, junsun]@sjtu.edu.cn.}
}


%


\maketitle

\begin{abstract}
Scene text recognition plays an important role in many computer vision applications. The small size of available public available scene text datasets is the main challenge when training a text recognition CNN model. In this paper, we propose a CNN based Chinese text recognition algorithm. To enlarge the dataset for training the CNN model, we design a synthetic data engine for Chinese scene character generation, which generates representative character images according to the fonts use frequency of Chinese texts. As the Chinese text is more complex, the English text recognition CNN architecture is modified for Chinese text. To ensure the small size nature character dataset and the large size artificial character dataset are comparable in training, the CNN model are trained progressively. The proposed Chinese text recognition algorithm is evaluated with two Chinese text datasets. The algorithm achieves better recognize accuracy compared to the baseline methods.
\end{abstract}


%
\IEEEpeerreviewmaketitle

\section{Introduction}
Extracting text information from natural images is conducive to wild areas of computer vision.
Generally, text information extraction is divided into two stages: text detection and text recognition, for efficiency. As text recognition in scanned documents is well studied, most existing researches focus on the text detection stage. However, when applied to scene text recognition, which meets challenges such as complex backgrounds and geometric distortions, the performance of traditional OCR techniques drops sharply. Therefore, recently there are more and more text recognition algorithms reported in the literatures~\cite{Goodfellow2013, Neumann2012, Yao2014}. To quantify and track the progress of text detection and recognition in natural images, several competitions, including the ICDAR robust reading competitions in 2011, 2013 and 2015~\cite{ICDAR2011,ICDAR2013,ICDAR2015}, have been held in recent years.

Recently, deep learning models are wildly used in text recognition algorithms. The work in~\cite{Jaderberg2014} empirically studied some convolutional neural network (CNN) models for English word recognition. They design a synthetic data engine to generate scene word images artificially to train the CNNs because of the small size of existing text recognition datasets.

It is noted that the above reported works are mainly focused on English text recognition, while few research works on Chinese text recognition have been reported. Chinese characters are more complex than English characters in both number and types of strokes. Therefore, the complexity of Chinese characters requires the recognition algorithms to focus more on characters rather than words in English text recognition.

In this paper, we make two major contributions. First, we design a synthetic data engine for Chinese scene character generation. The majority of Chinese fonts are derived from a few basic fonts. Based on this phenomenon, the Chinese synthetic data engine controls the proportion of generated characters with different fonts in order to reduce the training computational complexity with certified recognition accuracy. Second, we propose a state-of-the-art CNN Chinese scene text recognizer that trained by the mixture of artificial and nature character images. The CNN architecture for English text recognition is not capable to extract Chinese text features as the convolutional window is too small for the complex text strokes. We enlarge the window size of English text recognition CNN to extract Chinese text features more efficiently.
In order to ensure the nature and artificial character datasets, which have huge difference in size, are comparable in training, the CNN model are trained progressively.

\section{Proposed Algorithm}

\subsection{Synthetic Data Engine}

To train a CNN to recognize all the Chinese characters used nowadays, which is counted more than 10,000, the size of training dataset is demanded to be millions. Even for the most common used  1200 Chinese characters, the size of training dataset is also demanded to be several hundreds of thousands. However, there is few public available Chinese text recognition datasets~\cite{Ren2015, Zhou2015}, the number of Chinese scene characters in only in thousands, and the vocabulary is very limited.

We follow some success synthetic character datasets and design a Chinese synthetic character generator. The generation process of the  synthetic data engine are: First, the base characters are generated. The font is randomly selected from 32 Chinese fonts, which includes three basic fonts Kai, Song and Hei, 24 derived fonts such as FangSong, XiHei and 5 common special fonts. When generating characters, 30\% of them select basic fonts, 60\% select derived fonts and 10\% select special fonts. The character color and background color are randomly selected form the most common 27 colors in scene text images. Second, the base characters are processed into scene characters. Borders or shadows are added randomly to the characters. Then full-projective transformation is used to distort the characters. Finally, the scene characters are processed into scene character images. Scene background patches are blended to the characters to imitate the reflection in real word. Then Gaussian noise and Gaussian blur are added to the characters with random intensity.

\subsection{The CNN model}

The CNN Chinese text recognition model has 7 layers including three convolutional layers, two max-pooling down-sampling layers and two fully connected layers. The down-sampling layers connected after the second and third convolutional layers.

The CNN model are trained in two steps. In the first step, the training set only includes artificial character images. While in the second step, the artificial character images and the nature character images of the training set are approximately equal in quantity because the vocabulary of nature character images are limited.

\section{Experiments}

\begin{figure}[!tbp]
  \centering
  \includegraphics[width=0.15\linewidth]{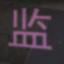}
  \includegraphics[width=0.15\linewidth]{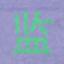}
  \includegraphics[width=0.15\linewidth]{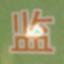}
  \includegraphics[width=0.15\linewidth]{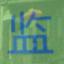}
  \includegraphics[width=0.15\linewidth]{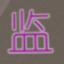}
  \includegraphics[width=0.15\linewidth]{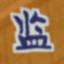}

  \caption{ Artificial character image examples}
  \label{artex}
\end{figure}

Our Chinese text recognition algorithm is evaluated on two text recognition datasets. One of them is an all natural text image dataset~\cite{Ren2015} in which all the text images are taken from natural. The other dataset~\cite{Zhou2015} is a partly natural text image dataset which contains a number of artificial text images like Internet ADs.

In order to train a credible CNN recognizer, approximately 200,000 artificial Chinese character images (Some examples are shown on Fig.1) are generated by the synthetic data engine described in Section II.A. As the Chinese character images in both of the training sets (about 4000 in the all scene text image dataset and about 8000 in the partly scene text image dataset)  are very little compare to the artificial character images, the CNN model are trained in two steps to ensure the artificial training set and scene training set are comparable. In the first step, 95\% artificial Chinese character images and no scene images are used to train the CNN model. And in the second step, the CNN model are training by the combination of all the scene training images and the remaining artificial images.

\begin{table}[!h]
\centering
\caption{\textbf{Evaluation results of different methods}}\label{rres}
\begin{tabular}{|c|c|c|c|}
    \hline
    \multicolumn{2}{|c|}{\multirow{2}{*}{Model}} & \multicolumn{2}{|c|}{Accuracy}\\
    \cline{3-4}
    \multicolumn{2}{|c|}{} & Ren's dataset~\cite{Ren2015} & Zhou's dataset~\cite{Zhou2015}\\
    \hline
    \multirow{3}{*}{CNN-7} & A & 0.42 & 0.70 \\
    \cline{2-4}
    & S & 0.38 & 0.45 \\
    \cline{2-4}
    & A + S & 0.73 & 0.76 \\
    \hline
    \multirow{3}{*}{CNN-9} & A & 0.44 & 0.69 \\
    \cline{2-4}
    & S & 0.31 & 0.36 \\
    \cline{2-4}
    & A + S & 0.73 & 0.75 \\
    \hline
    \multicolumn{2}{|c|}{ABBYY} & 0.40 & 0.41 \\
    \hline
\end{tabular}

\end{table}

Table I summarizes the evaluation results of different Chinese text recognition methods in both of the text recognition datasets. ``A'' presents the artificial character image dataset generated by synthetic data engine, and ``S'' presents the scene character image dataset extracted from the respective training sets. CNN-9 is a 9-layer CNN model following the text recognition CNN model in ~\cite{Jaderberg2014}, which have more layers than our proposed CNN model. It can be noted that although the CNN-9 is deeper, the performance is worse than our proposed CNN model. We believe that the relationships between Chinese strokes are more complex than English strokes. Thus a filter with smaller size, which is usually designed for deeper networks, is not suitable to recognize the complex relationships between Chinese strokes.

\section{Conclusion}

In this paper, we present a scene Chinese text recognition algorithm based on CNN model.  A synthetic data engine, which generates representative artificial character images according to the fonts use frequency of Chinese texts, is designed to enlarge the dataset for training the CNN model. The CNN model are trained in two steps to ensure the small size nature character dataset and the large size artificial character dataset are comparable in training. Our experimental results demonstrated that the proposed algorithm is effective in Chinese scene text recognition in several respects: 1) the artificial is helpful in training CNN; 2) deeper network shows no advantage in recognizing Chinese characters because its filter size is smaller.


\section*{Acknowledgment}

The work is partially supported by the National Natural Science Foundation of China(Grant No. 61201384, 61221001) and Shanghai Science and Technology Committees of Scientific Research Project(Grant No. 14DZ1101200).



%
\nocite{ICDAR2011,ICDAR2013,ICDAR2015,Ren2015,Jaderberg2014,Goodfellow2013,Zhou2015,Neumann2012,Yao2014}
\bibliography{bare_conf}
\bibliographystyle{IEEEtran}

\end{document}